\documentclass[conference]{IEEEtran}
\IEEEoverridecommandlockouts
\usepackage{amsmath}
\usepackage{amssymb}
\usepackage{amsfonts}

\usepackage{graphicx}

\usepackage{booktabs}
\usepackage{array}

\usepackage{cite}

\usepackage[hidelinks]{hyperref}

\usepackage{algorithm}
\usepackage{algpseudocode}

\usepackage{tikz}
\usetikzlibrary{
    arrows.meta,
    positioning,
    shapes,
    fit
}

\usepackage{enumitem}
\makeatletter
\newcommand{\linebreakand}{%
  \end{@IEEEauthorhalign}
  \hfill\mbox{}\par
  \mbox{}\hfill\begin{@IEEEauthorhalign}}
\makeatother

\usepackage{xcolor}
\newcommand{\redcircle}[1]{%
  \tikz[baseline=-0.6ex]{
    \node[
      circle,
      fill=red!85!black,
      text=white,
      font=\large\bfseries,
      inner sep=0pt,
      minimum size=1.15em
    ] {#1};
  }%
}

\definecolor{demoGreen}{RGB}{82,145,44}

\newcommand{\greencircle}[1]{%
  \tikz[baseline=-0.6ex]{
    \node[
      circle,
      fill=demoGreen,
      draw=demoGreen,
      text=white,
      font=\large\bfseries,
      inner sep=0pt,
      minimum size=1.15em
    ] {#1};
  }%
}

\title{

UniDataAgent: An Ontology-Grounded Agent for Enterprise Question-to-Report Automation

}

\author{
\IEEEauthorblockN{Yutai Duan\textsuperscript{\textdagger}}
\IEEEauthorblockA{
\textit{China Unicom Software}\\
\textit{Research Institute}\\
Beijing, China \\
yutaiduan@gmail.com}\\

\and
\IEEEauthorblockN{Yahui Zhao\textsuperscript{\textdagger}}
\IEEEauthorblockA{
\textit{China Unicom Software}\\
\textit{Research Institute}\\
Beijing, China \\
zhaoyh99@chinaunicom.cn}
\and
\IEEEauthorblockN{Zhangti Li\textsuperscript{*}}
\IEEEauthorblockA{
\textit{China Unicom Software}\\
\textit{Research Institute}\\
Beijing, China \\
lizt21@chinaunicom.cn}
\and
\IEEEauthorblockN{Yu Ma}
\IEEEauthorblockA{
\textit{China Unicom Software}\\
\textit{Research Institute}\\
Beijing, China \\
may50@chinaunicom.cn}
\and
\linebreakand
\IEEEauthorblockN{Zhenfeng Qi}
\IEEEauthorblockA{
\textit{China Unicom Software}\\
\textit{Research Institute}\\
Beijing, China \\
qizf8@chinaunicom.cn}
\and
\IEEEauthorblockN{Shaoyang Yuan}
\IEEEauthorblockA{
\textit{China Unicom Software}\\
\textit{Research Institute}\\
Beijing, China \\
yuansy39@chinaunicom.cn}
\and
\IEEEauthorblockN{Jing Fan}
\IEEEauthorblockA{
\textit{China Unicom Software}\\
\textit{Research Institute}\\
Beijing, China \\
fanj26@chinaunicom.cn}

\and
\IEEEauthorblockN{Jie Liu}
\IEEEauthorblockA{
\textit{College Of Artificial Intelligence,}\\
\textit{Nankai University}\\
Tianjin, China \\
jliu@nankai.edu.cn}
\thanks{\textdagger These authors contributed equally to this work.}
\thanks{*Corresponding author. Email: lizt21@chinaunicom.cn}
\thanks{This paper has been accepted by ICDM 2026 Demo track.}
}

\begin{document}
\maketitle

\begin{abstract}
Enterprise data agents must preserve organization specific semantics, not just translate questions into queries. We present ChinaUnicom DataAgent (UniDataAgent), an ontology grounded system for reusable question-to-report analysis that separates semantic acquisition from online execution. Ontology Acquisition and Validation stage (OAV) builds versioned enterprise ontologies from metadata, business knowledge, and supporting materials through expert authored business skills, constrained generation, question verification, and selected expert review. Question-to-Report Execution (QRE) stage retrieves semantic contracts for each question, coordinates skills and data tools, validates results, and produces evidence linked reports. Across 27 enterprise tables and roughly thousands of metric types, ontology construction took a few hours instead of about one week manually. It took just a few minutes to generate the reports, instead of several working days. Ontology grounding achieved 95.0\% strict accuracy on real business questions, versus 72.5\% for document RAG, especially on structured and compositional tasks. The system has already been deployed to generate cost savings and has the potential to be replicated in other enterprises. 

\end{abstract}


\section{Introduction}

Enterprise managers consume data via reports, not direct database queries. Even a simple request (e.g., quarterly broadband performance by region) requires cross-role scoping, authoritative calculation, and report preparation. In large organizations, this process takes days and often drifts from the original intent. The core challenge is preserving end-to-end business semantics, not merely generating SQL.

\begin{figure*}[t]\centering\includegraphics[width=0.90\textwidth]{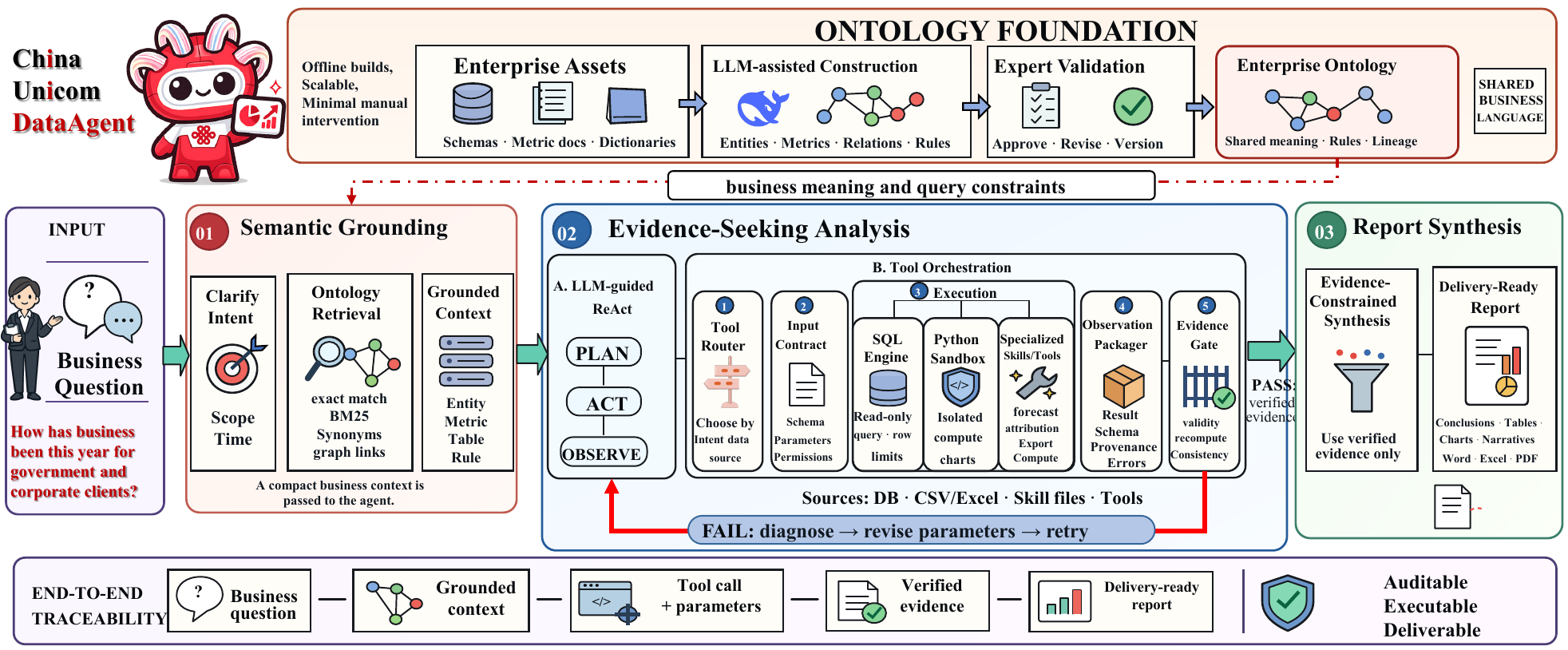}\caption{Overview of UniDataAgent. The previously defined OAV stage constructs and publishes a reusable ontology contract; the QRE stage repeatedly consumes question-specific ontology context, expert-defined analytical Skills, and executable evidence to produce delivery-ready reports.}\label{fig:overview}\end{figure*}

Recent systems only improve isolated steps: Text-to-SQL maps questions to executable queries~\cite{spider,bird}, ontology-based querying links language to logical and physical schemas~\cite{athena,queryartisan}, and LLM agents support reasoning and multi-step analysis~\cite{react,dabench,dbgpt}. Yet valid queries can still misalign on metrics, caliber, or scope, while plausible plans may violate schema-absent enterprise procedures. These gaps show question-to-report automation demands more than standalone query generation or tool use.

End-to-end question-to-report automation for enterprises requires two coupled capabilities: a reusable knowledge foundation and an agent that operationalizes it during analysis. Enterprise ontologies can provide the foundation by recording business concepts, metric definitions, relations, mappings, and provenance, but constructing them from fragmented metadata, documents, and expert knowledge is costly. Conversely, an LLM-based agent can coordinate reasoning and tools, but without an explicit ontology it must repeatedly infer the same business knowledge and cannot consistently apply it across questions. Our key principle is therefore to acquire, verify, and persist a reusable enterprise ontology, then selectively retrieve and execute it through an online data agent.

This perspective leads to three connected challenges. \textbf{C1: Acquiring enterprise semantics at scale.} How can fragmented metadata and business knowledge be compiled into a trustworthy, versioned ontology without requiring experts to manually author every entity, rule, and physical mapping? The construction process must control what may be generated, preserve provenance, test whether the resulting ontology supports actual business questions, and repair detected failures. \textbf{C2: Operationalizing expert analysis.} Once a question has been semantically grounded, how can the system execute organization-specific analytical procedures rather than merely produce plausible queries? The agent must select and parameterize expert-defined analytical Skills, coordinate them with data tools, and detect invalid units, periods, scopes, or intermediate results. \textbf{C3: Closing the evidence-to-report loop.} How can multiple tool observations be transformed into a report that remains traceable and suitable for analyst review? The system must preserve metric caliber and provenance, distinguish accepted evidence from execution artifacts, and constrain factual report content to executed observations.

To address these challenges, we present UniDataAgent, an ontology-grounded enterprise question-to-report system with two coupled stages. The offline Ontology Acquisition and Validation (OAV) stage interprets fragmented enterprise materials through expert-authored business skills, plans the semantic structure, generates ontology items under explicit type, mapping, and provenance constraints, and verifies them against representative business questions. Failures trigger localized repair or structural replanning, while uncertain or high-impact decisions and final release require expert review.
The online Question-to-Report Execution (QRE) stage clarifies analytical intent and retrieves a question-specific ontology contract encoding metric definitions, dimensions, physical mappings, calculation prerequisites, query constraints, and lineage. It combines this contract with analytical skills and executable tools, validates observations against business scope and caliber, and organizes accepted evidence into analyst-ready report artifacts. The stages interact through the typed ontology contract rather than an open natural-language handoff.

The demonstration of our system highlights three capabilities: (1) constructing and inspecting a reusable enterprise ontology from fragmented metadata and business materials; (2) grounding an ambiguous business question in executable metric definitions, source mappings, and query constraints; and (3) producing a delivery-ready report with charts, findings, and evidence-backed recommendations. The offline OAV stage makes enterprise semantics reusable across requests, while the online QRE stage turns the retrieved semantics into a validated analytical workflow.Across 27 enterprise tables and roughly thousands of metric types, ontology construction took a few hours instead of about one week manually. A 40-question comparison shows that the largest accuracy gains over document RAG occur on tasks requiring set-level structure or alignment across multiple business metrics.

\section{System Overview}

UniDataAgent addresses the three challenges through two coupled stages (Fig.~\ref{fig:overview}). OAV transforms fragmented enterprise materials into a versioned, reusable ontology (C1). QRE consumes this ontology for question-specific online analysis and evidence-linked report delivery (C2--C3). An explicit ontology contract makes offline knowledge directly executable online. A short demonstration video is \href{https://youtu.be/khqt5gf1ZvI}{\textcolor{red}{available online}} (YouTube).

\subsection{OAV: Constructing the Semantic Foundation}

OAV addresses C1 through a four-stage construction loop. (1) It interprets database metadata, business knowledge, and supplementary materials with an expert-authored Business Skill, which normalizes terminology, metric calibers, organizational relations, source priorities, and admissible mappings. (2) It converts these inputs into a semantic plan covering business domains, entity and relation types, report classes, physical mappings, provenance requirements, and prohibited content. The plan is checked for missing coverage, conflicting definitions, unsupported mappings, and insufficient source support before candidate generat

(3) OAV generates ontology candidates under the approved plan and links each entity, metric, rule, relation, and mapping to its supporting source. (4) It constructs a knowledge-grounded verifier agent from the Business Skill, source constraints, and representative business questions. The verifier checks coverage, consistency, provenance, mappings, and question answerability, and triggers localized repair or replanning when needed. Candidates that pass verification are released as versioned ontologies. Domain-expert approval is recommended for unresolved or high-impact changes, allowing experts to review consequential decisions rather than manually author every ontology item.

\begin{figure}[t]
    \centering
    \includegraphics[width=\linewidth]{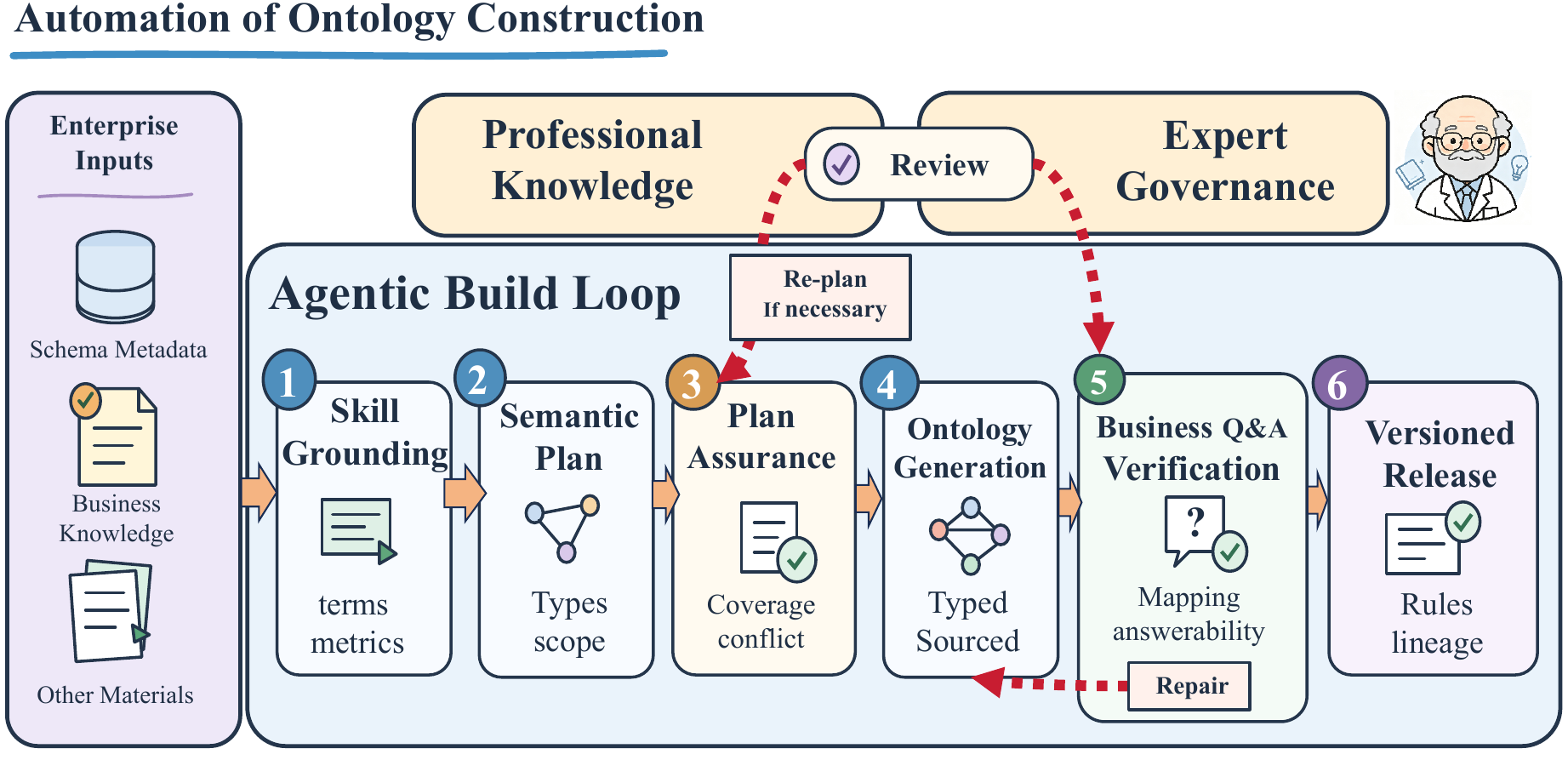}
    \caption{OAV constructs the enterprise ontology through input interpretation, Business-Skill grounding, semantic planning, constrained generation, business-question verification, and expert-governed release. Verification failures trigger localized repair or structural replanning.}
    \label{fig:OAV}
\end{figure}

\subsection{QRE: From Questions to Evidence-Backed Reports}

QRE leverages the published ontology to address challenges C2 and C3.
(1) It first identifies core query attributes: entity, metric, population, organizational and geographic scope, time range, comparison basis, analytical objective, and output form. It requests clarification only when required business parameters are missing or ambiguous.
(2) It retrieves compact ontology context via exact and alias matching, lexical retrieval, synonym expansion, and ontology relations. This context maps user expressions to canonical entities and metrics, and provides dimensions, physical fields, aggregation rules, metric calibers, query constraints, and lineage.
Relevant business documents and procedural rules serve as supplementary constraints, with the ontology as the primary enterprise knowledge source.

(3) QRE then runs a ReAct-style \emph{plan--act--observe} loop. It first plans analytical subgoals and selects skills or tools (e.g., read-only SQL, sandboxed computation) for data retrieval or computation. Each observation is validated against metric caliber, units, scope, time period, provenance, and prior results. Failed observations trigger revision, cross-source verification, recomputation, or replanning.
(4) Once sufficient evidence is gathered, the LLM synthesizes key findings, rankings, changes, anomalies, trends, and evidence-backed operational suggestions from a structured evidence package. The front end renders results into tables and charts, producing reports with data views, diagnostic analysis, caliber definitions, actionable insights, and evidence citations. Factual claims are anchored to executed observations, while interpretations and recommendations are explicitly marked as evidence-supported analysis.

\section{Demonstration}

\begin{figure}[t]
    \centering
    \includegraphics[width=0.9\linewidth]{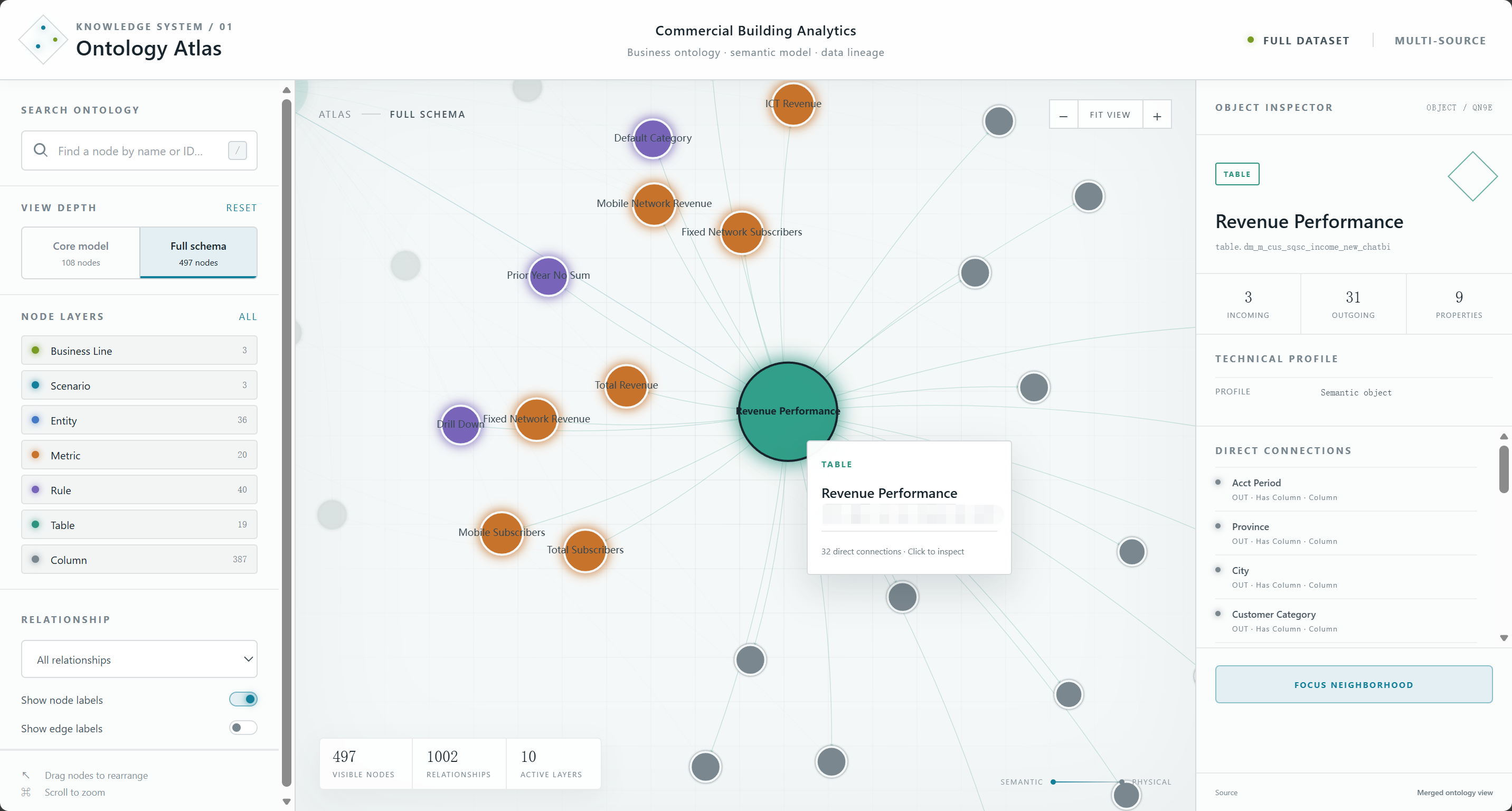}
    \caption{The enterprise ontology is automatically constructed, organizing hundreds of domain concepts and data objects and thousands of semantic relations into an inspectable knowledge structure in a few hours.}
    \label{fig:ontology}
\end{figure}

\begin{figure}[t]
    \centering
    \includegraphics[width=\linewidth]{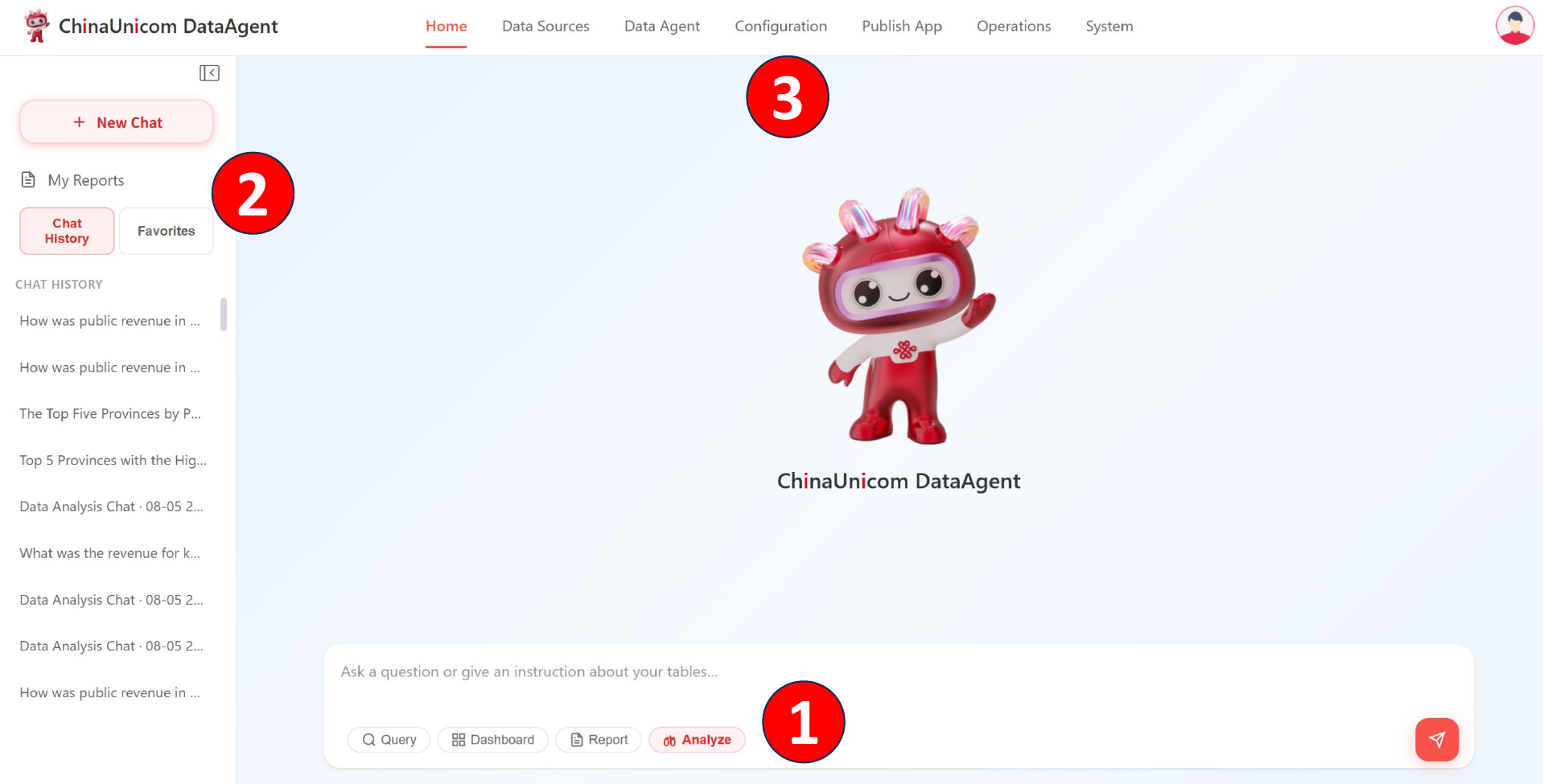}
    \caption{Main interface of UniDataAgent. The numbered regions indicate the question-and-mode workspace, session controls, and navigation functions.}
    \label{fig:demo}
\end{figure}
\begin{figure}[t]
    \centering
    \includegraphics[width=\linewidth]{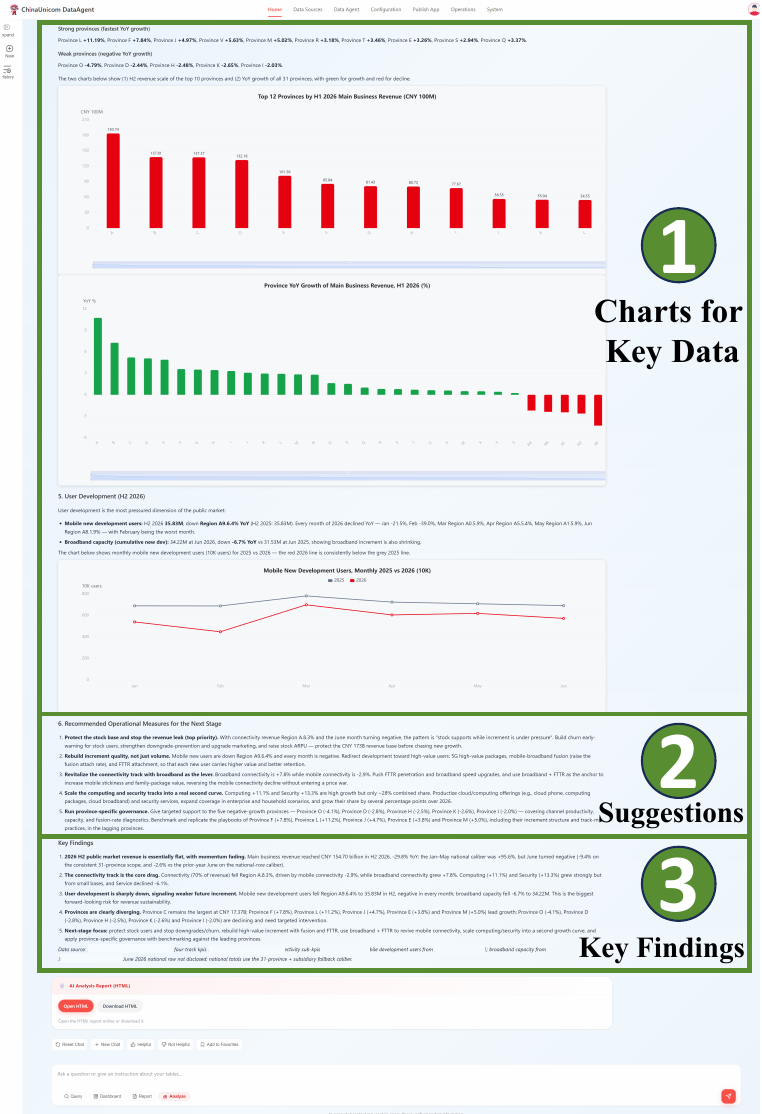}
\caption{
Partial answers to: ``How will the public market perform in 2026? Which provinces will perform well or be weaker? What next-step business strategies are recommended?'' The report includes evidence-linked charts, candidate suggestions for analyst review, and key findings grounded in retrieved results. All demonstration data in this figure were irreversibly transformed; no original numerical values, locations, or identifiers are included.
}
    \label{fig:case}
\end{figure}
Fig.~\ref{fig:ontology} presents the automatically constructed ontology. Within approximately a few hours, hundreds of domain concepts and data objects, together with thousands of semantic relations, are organized into an inspectable knowledge structure. The ontology provides explicit and traceable grounding for downstream agents by linking defined entities, metrics, rules, and data objects. Business users can inspect inter-entity relations and detailed semantic descriptions, allowing subsequent question answering to rely on defined business semantics rather than ungrounded inference. This inspection allows users to verify the semantic basis of an analysis before execution.

Building on this ontology foundation, Fig.~\ref{fig:demo} shows the demonstration interface. Users enter a question and select the required output mode in the central workspace (\redcircle{1} in Fig.~\ref{fig:demo}); the system then retrieves the relevant business context, including entities, metrics, and data constraints. The remaining controls provide session configuration (\redcircle{2} in Fig.~\ref{fig:demo}) and access to conversations, reports, data sources, agents, and administrative functions (\redcircle{3} in Fig.~\ref{fig:demo}). The demonstration focuses on enterprise databases, although uploaded tables are also supported.

Using the grounded context, Fig.~\ref{fig:case} presents an end-to-end case. Given a natural-language analytical request, the system retrieves task-relevant records under source, temporal, metric, and geographic-granularity constraints. It organizes revenue rankings, year-over-year changes, and user-development trends into revisitable charts (\greencircle{1} in Fig.~\ref{fig:case}), generates candidate suggestions from cross-period comparisons (\greencircle{2} in Fig.~\ref{fig:case}), and summarizes evidence-linked findings for analyst review (\greencircle{3} in Fig.~\ref{fig:case}). The demonstrated workflow completed in approximately 200 seconds, including data screening, cross-table reconciliation, chart generation, and report synthesis.

\section{Real-World Enterprise Question Evaluation}

\subsection{Evaluation Setup}We evaluate 40 real enterprise questions from six operational domains. Rather than grouping questions by reasoning depth, we assign each question to exactly one task type according to its primary semantic requirement: (1) \emph{metric and scope grounding} (22 questions), which requires resolving the correct metric, population, time window, or business category; (2) \emph{complete regional comparison} (11 questions), which requires exhaustive ranking, filtering, or counting across provinces; and (3) \emph{cross-metric business diagnosis} (7 questions), which requires jointly interpreting multiple indicators or their relationships. Each question is paired with a reference answer and a checklist covering the required result, scope, conditions, and metric caliber. We used DeepSeek-V4.

We compare UniDataAgent with document RAG in the same enterprise data environment. Document RAG retrieves relevant business documents but must infer metric definitions, aggregation rules, dimensions, and physical mappings at query time. UniDataAgent instead retrieves an ontology contract that connects business concepts with executable constraints.

\subsection{Results and Failure Analysis}

\begin{table}[t]
\centering
\caption{Strict answer accuracy by mutually exclusive task type.}
\label{tab}
\resizebox{\columnwidth}{!}{
\begin{tabular}{lccc}
\hline
Task Type & \#Q & Doc-RAG & Ours \\
\hline
Metric and scope grounding       & 22 & 81.8\% & 90.9\% \\
Complete regional comparison     & 11 & 63.6\% & 100.0\% \\
Cross-metric business diagnosis  & 7  & 57.1\% & 100.0\% \\
\hline
Overall                          & 40 & 72.5\% & 95.0\% \\
\hline
\end{tabular}
}
\end{table}

As shown in Table~\ref{tab}, UniDataAgent improves performance across all three task types, with the largest gains on regional comparison and cross-metric diagnosis. Document RAG may retrieve relevant values but still omit required provinces, apply inconsistent metric calibers, or form unsupported relations between indicators. In contrast, the ontology contract specifies valid mappings, scopes, and calculation constraints before execution, while evidence checks constrain the final analysis to supported observations. These deployment-specific results indicate that explicit ontology grounding is particularly valuable when a question requires complete coverage or business-consistent interpretation beyond isolated facts.

\section{Conclusion}

UniDataAgent implements an enterprise Question‑to‑Report workflow built on a persistent, reviewed semantic foundation. It cuts ontology‑construction costs and closes the report loop via offline semantic acquisition and online skill‑driven execution, generating traceable outputs with evidence‑constrained synthesis. Real‑world deployment and comparative tests demonstrate less semantic rework, higher analytical accuracy, lower delivery costs, and cross‑enterprise reuse potential.

\end{document}